\documentclass[10pt,twocolumn,letterpaper]{article}

\usepackage{wacv}
\usepackage[accsupp]{axessibility}
\usepackage{times}
\usepackage{epsfig}
\usepackage{graphicx}
\usepackage{amsmath}
\usepackage{amssymb}
\usepackage{multirow}
\usepackage{epsfig}
\usepackage{comment}
\usepackage[table,dvipsnames]{xcolor}
\usepackage{enumitem}
\usepackage{array}
\usepackage{varwidth}
\usepackage[ruled,vlined]{algorithm2e}
\usepackage{cases}
\usepackage{amsthm}
\usepackage{subcaption}
\usepackage{booktabs}
\usepackage{mathtools}
\usepackage{amsmath}
\usepackage{multicol}
\usepackage[export]{adjustbox}
\usepackage[accsupp]{axessibility}
\usepackage{pdfpages}
\usepackage{multido}

\usepackage[pagebackref=true,breaklinks=true,letterpaper=true,colorlinks,bookmarks=false]{hyperref}
\usepackage[capitalise]{cleveref}
\Crefname{table}{Tab.}{Figs.}
\Crefname{section}{Sec.}{Figs.}

\newcommand{\gtacs}{GTA5$\rightarrow$Cityscapes}
\newcommand{\synthiacs}{SYNTHIA$\rightarrow$Cityscapes}

\def\wacvPaperID{434} 

\wacvfinalcopy 

\ifwacvfinal
\fi

\ifwacvfinal
\usepackage[breaklinks=true,bookmarks=false]{hyperref}
\else
\usepackage[pagebackref=true,breaklinks=true,colorlinks,bookmarks=false]{hyperref}
\fi

\ifwacvfinal
\else
\fi

\begin{document}

\title{Shallow Features Guide Unsupervised Domain Adaptation for Semantic Segmentation at Class Boundaries}

\author{Adriano Cardace\qquad Pierluigi Zama Ramirez\qquad Samuele Salti\qquad Luigi Di Stefano\\
Department of Computer Science and Engineering (DISI)\\
University of Bologna, Italy\\
{\tt\small \{adriano.cardace2,  pierluigi.zama\}@unibo.it}
}

\maketitle
\thispagestyle{empty}

\begin{abstract}
Although deep neural networks have achieved remarkable results for the task of semantic segmentation, they usually fail to generalize towards new domains, especially when performing synthetic-to-real adaptation.
Such domain shift is particularly noticeable along class boundaries, invalidating one of the main goals of semantic segmentation that consists in obtaining sharp segmentation masks.

In this work, we specifically address this core problem in the context of Unsupervised Domain Adaptation and present a novel low-level adaptation strategy that allows us to obtain sharp predictions. 
Moreover, inspired by recent self-training techniques, we introduce an effective data augmentation that alleviates the noise typically present at semantic boundaries when employing pseudo-labels for self-training.
Our contributions can be easily integrated into other popular adaptation frameworks, and extensive experiments show that they effectively improve performance along class boundaries. 
\end{abstract}

\begin{figure}[t]
    \centering
    \includegraphics[width=1\linewidth, trim={4.6cm 1cm 4.5cm 1cm}, clip,]{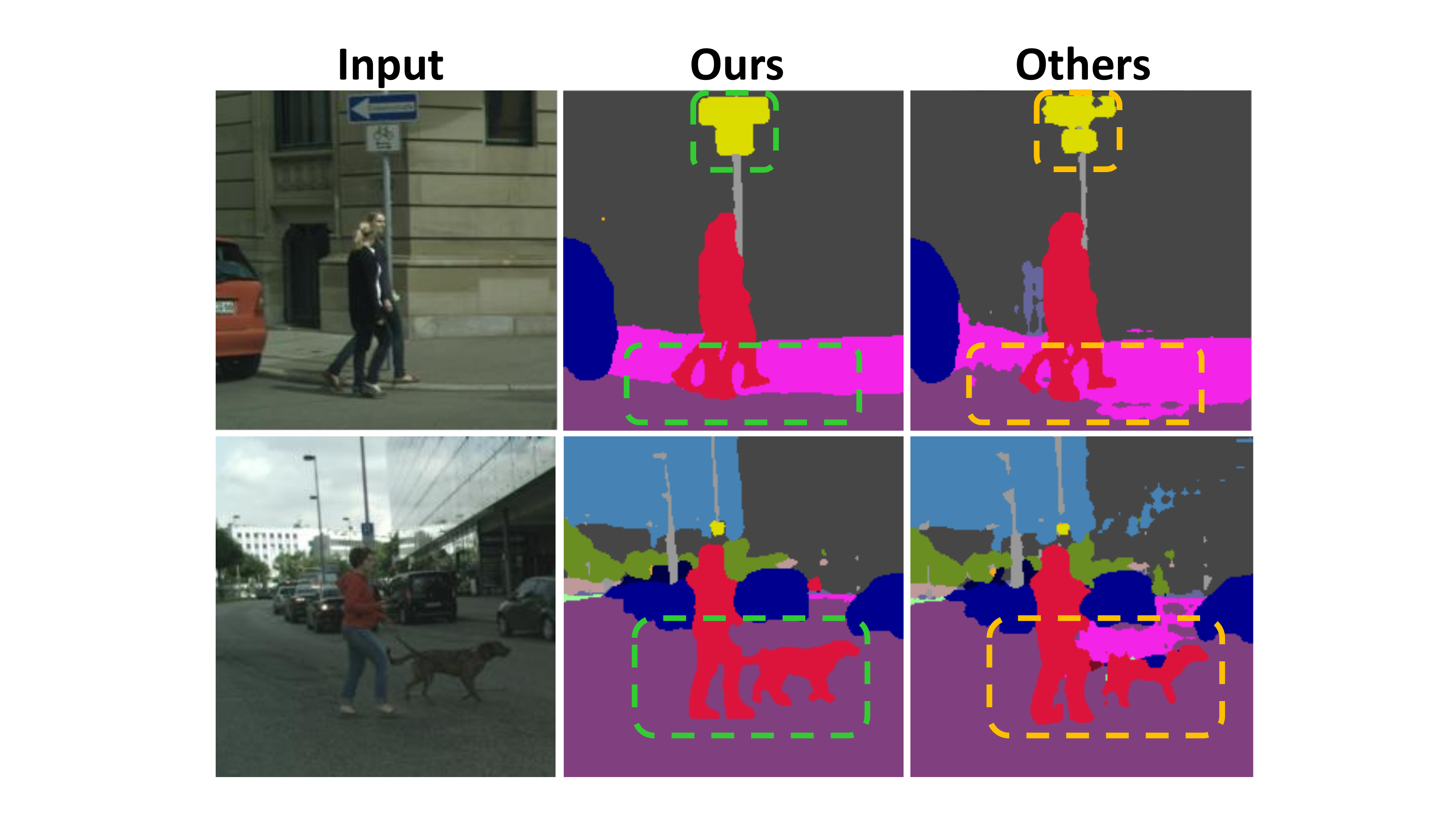}
    \caption{Given in input an RGB image (left-most column), our model produces sharp predictions along class boundaries (central column), while a model trained on translated images (right-most column) exhibits severe noise.}
    \label{fig:teaser}
\end{figure}

\section{Introduction}

Semantic segmentation is the process of assigning a class to each pixel of an image.
Recently, convolutional neural networks have proven to be highly effective in solving this challenging visual task \cite{cnnss, deeplabv2, Enet, unet}, leading to ever-increasing interest in the deployment of semantic segmentation models in spaces as diverse as autonomous driving, robotics, and medicine. 
However, training a semantic segmentation network requires a large amount of pixel-wise annotated data, which are tedious, time-consuming, and expensive to collect.
Moreover, current models often fail to generalize toward new domains, an issue that cannot be overlooked in many relevant real-world applications.
Indeed, performance often drops when models are tested on new scenarios, especially when there exists a domain gap between the training (source) and test (target) images. 
For instance, in autonomous driving settings, object appearances may drastically change when training and testing across different cities, leading to severe segmentation errors. This problem is even more pronounced when relying on synthetic data generated by computer graphics, such as video games \cite{gta} or 3D simulations \cite{synthia}, that could otherwise be advantageously exploited to easily obtain large amounts of labeled data.

Unsupervised Domain Adaptation (UDA) \cite{Wang_2018} aims at minimizing the impact of the domain gap under the assumption that no ground-truth annotations are available for the target domain. 
In the last few years, several UDA techniques have been proposed for the task of semantic segmentation \cite{hoffman2016fcns, adaptsegnet, ramirez2018exploiting, cycada, dcan, Chang_2019}. However, all these methods ignore the main goal of semantic segmentation, which is to obtain sharp prediction masks and only focus in the feature adaptation part. 
For this reason, previous works can correctly segment out coarse blobs of large elements in a scene such as cars or buildings, while they provide inaccurate segmentation masks along class boundaries as shown in \cref{fig:teaser}.

On the other hand, in the supervised semantic segmentation setting, a large amount of works focus on obtaining sharp predictions \cite{SEMEDA, Ke_2018, Chen_2016,Yuan_2020,Ding_2019}. This is commonly done by better integrating low-level features into high-level features since modern segmentation architectures discard spatial information with down-sampling operations such as max-pooling or strided convolution due to memory and time constraints.
Following the supervised setting, we argue that this line of research should also be pursued for the UDA case to obtain sharp predictions across domains even though target labels are not available.
Our approach, also leverages on low-level features to seek this goal, and we introduce a novel low-level adaptation strategy specifically for the UDA scenario.
More precisely, we enforce alignment of low-level features exploiting an auxiliary task that can be solved for both domains in a self-supervised fashion, intending to make them more transferable. By doing this, we enable the possibility to exploit shallow features to refine the coarse segmentation masks for both the source and target domains.
To achieve this, we estimate a 2D displacement field from the aligned shallow features that, for each spatial location of the predicted coarse feature map, specifies the direction where the representation for that patch is less ambiguous (i.e. at centre of the semantic object). Our intuition is that when the coarse feature map is bi-linearly up-sampled to regain the target resolution, the feature representation of those patches corresponding to semantic boundaries in the input image is mixed up, as it contains semantic information belonging to different classes.
Thanks to the estimated 2D displacement field, however, we refine each patch representation according to the features coming from the center of the object, which are less prone to be influenced by other classes as they lay spatially far from boundaries.
This process will be referred later as the feature \textit{warping} process.

Finally, following a recent trend in UDA for semantic segmentation \cite{iast, mrnet, Zheng_2021, Paul_WeakSegDA_ECCV20}, we employ self-training, a technique that foresees the training of a neural network with its own predictions denoted as pseudo-labels. This step allows to implicitly encourage cross-domain feature alignment thanks to the simultaneous training on multiple domains. Yet, differently from previous works that mainly focus on masking incorrect pixels with some heuristics, we propose a novel data augmentation technique aimed at preserving information specifically along class boundaries. 
In fact, due to the low confidence of the network in the target domain, pixels along edges are usually masked by the aforementioned methods, resulting in a further performance degradation along class boundaries due to the lack of supervision during the self-training process. Thus, we employ a class-wise erosion filtering algorithm that allows us to synthesize new training samples in which only the inner body of the target objects is preserved and copied into other images. By doing this, all pixels have supervision, and the network is trained to classify correctly edges also in the target domain. Code available at {\small{\url{https://github.com/CVLAB-Unibo/Shallow_DA}}.}
To summarize our contributions are:
\begin{itemize}
    \item We propose to use shallow features to improve the accuracy of the network along class boundaries in the UDA scenario. This is achieved by computing a displacement field that lets the network use information from the center of semantic blobs.
    
    \item We deploy semantic edge detection as an auxiliary task to enforce the alignment of shallow features, which is key to overcome the domain shift when computing the displacement map.
    
    \item We introduce an effective data augmentation that selects objects from target images and filters out noise at class boundaries to obtain sharp pseudo-labels.
    
    \item We show that our approach achieves overall on par or even state-of-the-art performance in standard UDA for semantic segmentation benchmarks, and more importantly improves predictions along boundaries when compared to previous works.
\end{itemize}

\section{Related Work}

\subsection{Pixel-level Domain Adaptation}
Pixel-level adaptation aims at reducing the visual gap between source and target images. Typically, style and colors are adapted by deploying CycleGANs\cite{cyclegan}, a generative model able to capture the target style and injecting it into the source images without altering their content. Early works \cite{dcan, cycada} learn such transformation offline, and employ the translated images during training time. Recent approaches instead~\cite{bdl, stylization}, fuse the translation process into the training pipeline, obtaining an end-to-end framework. ~\cite{ltir} extended this approach to obtain a texture-invariant network by training on source images augmented with textures from other natural images.
Following recent works, our approach builds upon these techniques. Indeed, we make use of translated images to obtain strong baseline and extract good pseudo-labels when adapting from synthetic to target.

\subsection{Adversarial Learning}
The goal of adversarial training in the context of Domain Adaptation is to align the distributions of source and target images so that the same classifier can be seamlessly applied on a shared feature extractor. Adaptation can be forced either in feature space \cite{fada} or in output-space \cite{adaptsegnet}. Many extensions of \cite{adaptsegnet} have been introduced. \cite{stuffAndThings} proposed to align differently classes based on their intra-class variability in their appearance. Other works deploy adversarial learning to minimize the entropy of the target classifier \cite{advent} or to perform feature perturbation \cite{perturbation}. 
In our work, since training a network adversarially is notoriously a difficult and unstable process \cite{salimans2016improved}, we avoid it.

\subsection{Self-Training}
A recent line of research focuses on self-training~\cite{pseudolabel} thanks to its effectiveness and simplicity. This approach is based on the idea of producing pseudo-labels for the target domain and use them to capture domain-specific characteristics. \cite{cbst} proposes an algorithm to filter out wrong pixels with some confidence thresholds. Similarly, \cite{iast} extended the idea by introducing an instance adaptive algorithm to improve the quality of pseudo-label.
\cite{mrnet} proposes to use pseudo-labels to minimize the discrepancy between two classifiers, while \cite{Pan_2020} tries to minimize both the inter-domain and intra-domain gap with the support of the pseudo-labels.
Differently, \cite{dacs} synthesizes new training samples by embedding objects from source images into the target ones. 
Inspired by these recent trends, we adopt self-training to align shallow features and guide the warping process across domains. Differently from previous approaches, however, we synthesise new training pairs enriching images of both domains with target objects to improve segmentation quality on class boundaries.

\begin{figure}[t]
    \centering
    \includegraphics[trim={2cm, 0.3cm, 1cm, 0.5cm}, clip, width=1\linewidth]{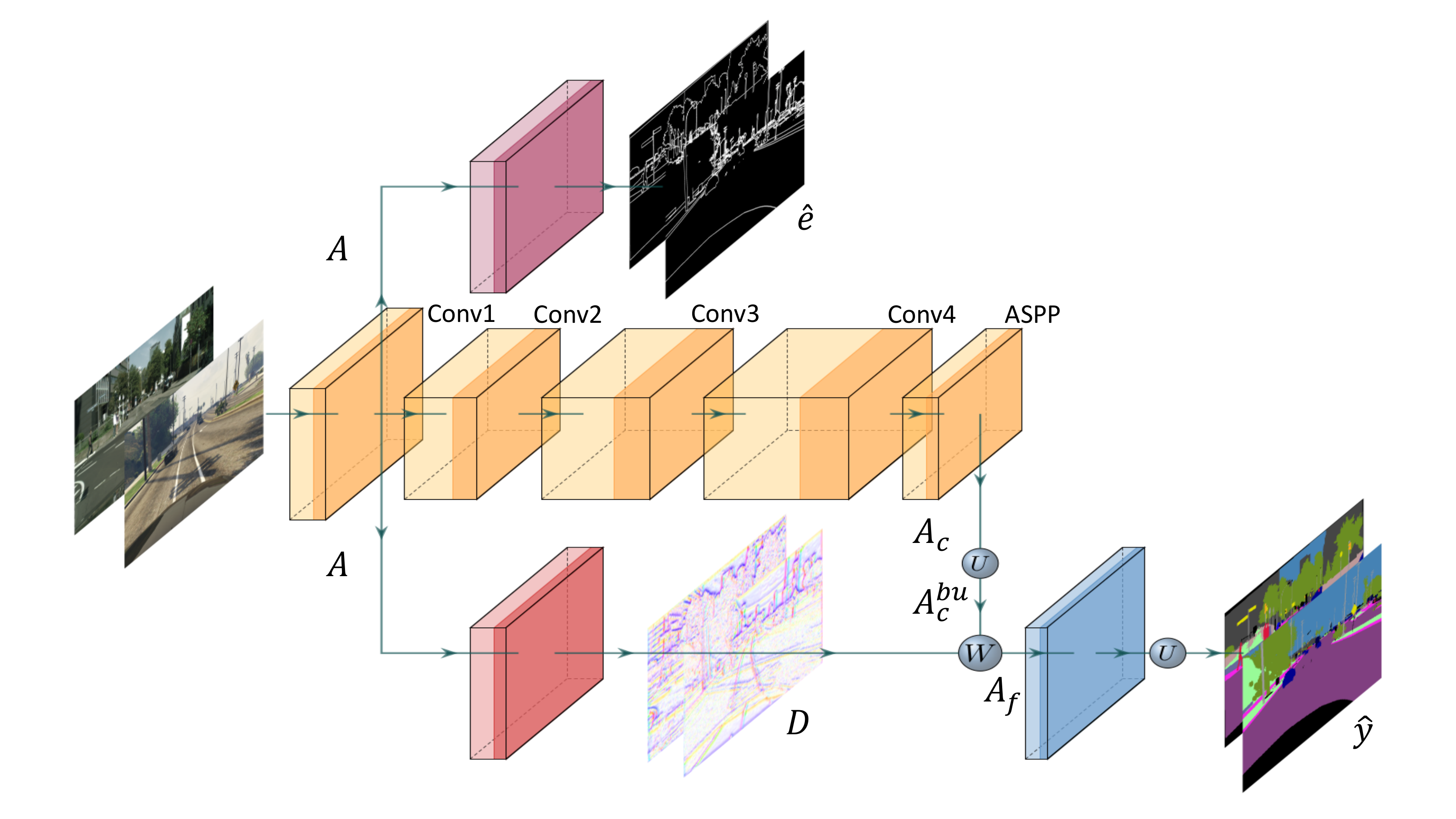}
    \caption{
    Illustration of our architecture in the adaptation step. Given an RGB input image, the network learns to extract semantic edges from shallow features. From the same feature map, a 2D displacement map is estimated in order to guide the warping of down-sampled deep features, which lacks of fine-grained details.}
    \label{fig:network}
\end{figure}

\section{Method}
In UDA for semantic segmentation we are given image-labels pairs $\{x_s^i, y_s^i\}_{i=1}^M$ for a source domain $\mathcal{S}$, while only images $\{x_t^i\}_{i=1}^N$ are available for a target domain $\mathcal{T}$. The goal consists in predicting pixel-wise classification masks for target images.
Our proposed framework comprises several components, as depicted in \cref{fig:network}. 
A standard backbone (yellow branch) produces a coarse feature map $A_c$ from an image.
A semantic edge extractor (top purple branch) estimates semantic edges $\hat{e}$, given the activation map $A$ produced by the first convolutional block of the backbone.
The same shallow features are processed by another convolutional block (bottom red branch) to obtain a 2D displacement map, $D$.
Then, $A_c$ is up-sampled to the same size as $D$ and it is refined according to $D$ to produce a fine-grained feature map $A_f$.
Finally, one last convolutional block that acts as a classifier is applied to produce a $C$-dimensional vector for each pixel, with $C$ being the number of classes, and a final bi-linear up-sampling yields a prediction map of the same size of the input.  
We detail each component in the following subsections.

\subsection{Low-level adaptation}
\textbf{Learning transferable shallow features}.
\label{sec:lowalign}
We introduce an auxiliary task to push the network to learn domain-invariant features that include details on objects boundaries already from early layers. 
Given the feature map $A$, a convolutional block $\gamma$ is applied to predict an edge map $\hat{e}$. 
Ground truths $e$ are obtained by the Canny edge detector \cite{canny} applied directly on semantic annotations for the source domain and on pseudo-labels for the target domain, so that only semantic boundaries are considered. A binary-cross entropy loss is minimized for batches including images from both domains:
\begin{equation}
\label{eq:edge_loss}
\begin{aligned}
\hat{e} &= \gamma(A), \\
\mathcal{L}_{edge} &= \sum_{h}^{H}\sum_{w}^{W} e^{(h,w)}\log \hat{e}^{(h,w)} \\
 &+ (1-e^{(h,w)})\log (1-\hat{e}^{(h,w)})
\end{aligned}
\end{equation}
Hence, we enforce the auxiliary semantic edge detection task for the very first layers of the network only, rather than, as in typical multi-task learning settings such as \cite{gebru2017fine, choi2020shuffle, sun2019unsupervised}, at a deeper level, where features are more task-dependent.
We believe this design choice to be key for a good generalization for three reasons.
First, trying to solve this task from shallow layers guides the network to explicitly reason about object shapes from the beginning, rather than solely texture and colors as typically done by CNNs \cite{textureVSshape}. 
Second, solving an auxiliary task for both domains forces the network to learn a shared feature representation, which naturally leads to aligned distributions.
Consequently, the displacement field generated from the shallow features is effective also in the target domain, and it can be directly exploited at a deeper level to recover fine-grained details.
Finally, the peculiar choice of semantic edge detection is directly beneficial to estimate a displacement field that mainly focuses on edges, making the following warping process more effective where the network is uncertain. We refer to the supplementary material for ablations on the alignment performed at different levels.

\textbf{Feature warping}.
One of the contributions of our method is to refine the bi-linearly up-sampled coarse feature map $A_c$, hereafter $A_c^{bu}$, to obtain a fine-grained feature map $A_f$ that better captures the correct class for pixels laying in the boundary regions. The refinement is guided by a 2D displacement field $D$ obtained from the domain-invariant shallow features computed by the first convolutional block of the backbone.
The displacement field indicates for each location of $A_c^{bu}$ where the network should look to recover the correct class information, namely the direction that better characterize that patch.
We estimate the 2D displacement map $D$ by applying a convolutional block to the aligned shallow features $A$ that are aligned as described above.

Our intuition is that, due to the unavoidable side-effect of the down-sample operations in the forward pass, the representation of those elements in $A_c$ whose receptive field includes regions at class boundaries in the original image, contains ambiguous semantic information. 
Indeed, when $A_c$ is bi-linearly up-sampled, patches that receive contributions from ambiguous coarse patches inherit such ambiguity. However, in the higher resolution feature map $A_c^{bu}$ it may be possible to compute a better, unambiguous representation for some of the patches, \ie those now laying entirely in a region belonging to one class.
The correct semantic information may be available in the nearby high-resolution patches closer to the semantic blob centers. Thus, each feature vector at position $p$ on a standard 2D spatial grid of $A_c^{bu}$, is mapped to a new position $\hat{p} = p + D(p)$, and we use a differentiable sampling mechanism \cite{stn} to approximate the new feature vector representation for that patch:

\begin{equation}
A_f(p) = \sum_{p_{l} \in \mathcal{N}(\hat{p})} w_{p_{l}} A_c^{bu}(p_{l})
\label{equ:warping}
\end{equation}

where $w_{p_{l}}$, are the bi-linear kernel weights obtained from $D$ and $\mathcal{N}$ the set of neighboring pixels. 
Hence, \cref{equ:warping} defines a backward warping operation in feature space, where $A_f$ is obtained by warping $A_c^{bu}$ according to $D$.
Finally, the fine-grained feature map $A_f$ is fed to the classifier to obtain the final prediction that is up-sampled by a factor of $2$ to regain the input image resolution. We minimize the cross entropy loss using annotations for the source domain and pseudo-labels for the target domain:

\begin{equation}
\label{eq:semantic_loss}
\mathcal{L}_{sem} = \sum_{h=1}^{H}\sum_{w=1}^{W} \sum_{c=1}^{C} y^{(h,w,c)} \log \hat{y}^{(h,w,c)}
\end{equation}

\subsection{Data Augmentation for Self-Training}
\label{sec:selftraining}
\begin{figure}[t]
    \centering
    \includegraphics[width=1\linewidth]{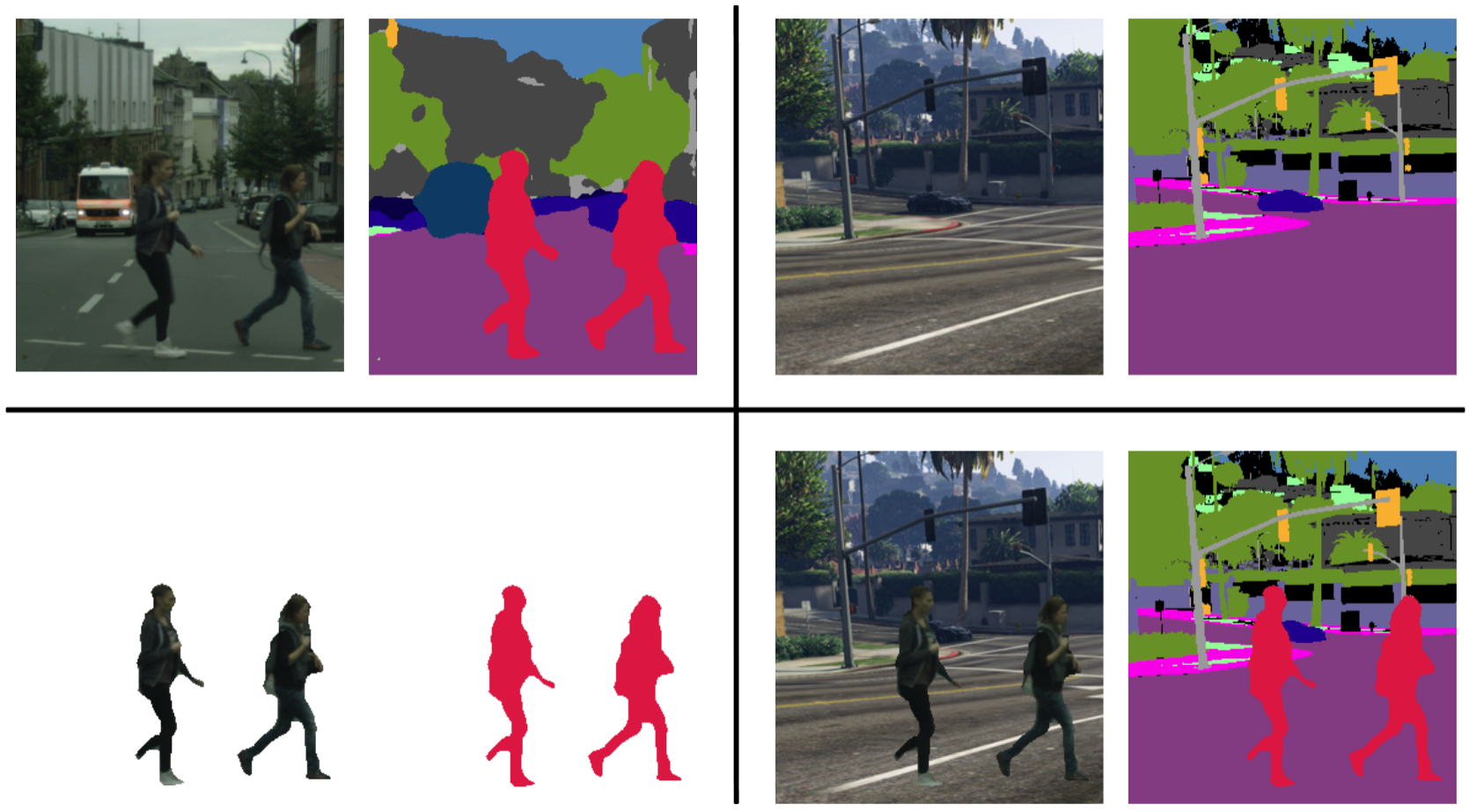}
    \caption{Given a target image prediction pair (top-left) and a source training pair (top-right), we select classes such as \textit{person} (bottom-left) and apply our class-wise data augmentation pipeline to synthesise a new training pair (bottom-right). The selected shapes are eroded before being pasted.}
    \label{fig:copypaste}
\end{figure}

Inspired by \cite{cutmix, Dwibedi_2017_ICCV, ghiasi2021simple, dacs}, we use a pre-trained model to select objects based on predictions on target images and paste them over source images (see \cref{fig:copypaste}).
Peculiarly, our self-training approach relies on a data augmentation process that selects objects from the target scenes rather than the source ones as done \cite{dacs}. Although selecting source objects may be useful to reduce the unbalanced  distributions of classes, it is a sub-optimal choice since the network would be still trained to identify shapes and details peculiar to the source domain, which are different to those found at inference time for the target images. We instead use pseudo-labels to cut objects from the target scenes and paste them into source or target images, forcing the network to look for these patterns on both domains.
However, due to the inherent noise of pseudo-labels we need to filter out noisy predictions. 
In particular, we aim at removing object boundaries as they typically exhibit classification errors and tend to be localized rather inaccurately.
Given a target image $x_t$ and its associated predictions $\hat{y}_t$, we compute a binary mask $B_c$ for each class $c \in C^*$, where $C^*$ denotes a random subset of the considered classes. 
We exclude classes such as \textit{"road"} and \textit{"building"} to avoid occlusion of the whole scene and to counteract the unbalanced distributions of classes, and only use object instances such as as \textit{"car"} and \textit{"poles"}.
This categorization is similar to the one used in \cite{stuffAndThings}, and can be easily adapted to different datasets. We refer to the supplementary material for the set of classes we used in each experiment.
For each spatial location \textit{p}, $B_c$ has value 1 if \textit{p} is assigned to class $c$, 0 otherwise. Then, we apply an erosion operation, $\ominus$, with a $5\times5$ structuring element $k$  to each class mask $B_c$. 
To obtain the set of pixels to be copied from the target image to a randomly selected source image we apply the union set operator to all masks:
\begin{align}
\label{eq:erosion}
B &= \bigcup_{c \in C^*} B_c \ominus k,
\end{align}
\begin{align}
x^p =
\begin{cases}
    x_t^p, & B^p=1 \\
    x_s^p, & B^p=0 
\end{cases},
y^p =
\begin{cases}
    \hat{y}_t^p, & B^p=1 \\
    y_s^p, & B^p=0
\end{cases}
\end{align}

The new synthesised training pairs are very often enriched with fine-grained details from the target domain. Indeed, as shown in \cref{fig:copypaste},  thanks to our data augmentation pipeline, only the inner part of an object is preserved while edges are discarded, producing sharp pseudo-labels even at class boundaries.
The whole data augmentation process is applied offline before training, therefore it does not have any impact on the training time.

\subsection{Training Procedure}
The whole pipeline can be summarised in 3 simple steps. We start with the \textit{initialization} step to train our baseline model (i.e. the yellow backbone of \cref{fig:network}) on the source domain only. We follow standard practices \cite{bdl, stuffAndThings, Tsai_2019, Paul_WeakSegDA_ECCV20, Yang_2021_WACV} and, for synthetic-to-real adaptation, we utilize domain-translated source images provided by~\cite{bdl}. 
We deploy this baseline to produce pseudo-labels for the target domain and obtain an augmented mixed dataset as detailed in \cref{sec:selftraining}.

Then, we perform the \textit{adaptation} step: we train the model illustrated in \cref{fig:network} that empowers our additional modules for low-level alignment as explained in \cref{sec:lowalign}.
It is important to highlight that the proposed data augmentation extracts objects from only target images and pastes them on images on both domains. Hence, at this stage, the training is done simultaneously on both domains.
The training loss is as follows:
\begin{equation}
\label{eq:finalloss}
\mathcal{L} = \mathcal{L}_{sem} + \lambda_{}\mathcal{L}_{edge}
\end{equation}
with $\lambda$ set to 0.1 in all experiments.

Finally, we use the predictions from the model trained in the previous step to synthesise new training pairs by following again the procedure detailed in \cref{sec:selftraining}. This allows us to distill the knowledge and the good precision along class boundaries of the previously enhanced model into a lighter segmentation architecture as the one used in the first step. We do this to avoid the introduction of additional modules at inference time.
Differently from the adaptation step however, we apply our data algorithm using solely images from the target domain. 
Indeed, as we are now at the third and final stage, we expect pseudo-labels to be less noisy compared to the previous step, 
and training only on the target domain allows to capture domain specific characteristic.
We denote this third step as the \textit{distillation} step.

\begin{table*}[ht!]
\begin{center}
\setlength{\tabcolsep}{2.5pt}
\scalebox{0.7}{
\begin{tabular}{l|cc|ccccccccccccccccccc|cc}
method & IT & ST &\rotatebox{90}{Road} & \rotatebox{90}{Sidewalk} & \rotatebox{90}{Building} &\rotatebox{90}{Walls} & \rotatebox{90}{Fence} & \rotatebox{90}{Pole} & \rotatebox{90}{T-light} & \rotatebox{90}{T-sign} & \rotatebox{90}{Vegetation} & \rotatebox{90}{Terrain} & \rotatebox{90}{Sky} & \rotatebox{90}{Person} 
& \rotatebox{90}{Rider} & \rotatebox{90}{Car}  & \rotatebox{90}{Truck} & \rotatebox{90}{Bus} & \rotatebox{90}{Train} & \rotatebox{90}{Motorbike} & \rotatebox{90}{Bicycle} & \textbf{mIoU}    \\    
\hline
AdaptSegNet~\cite{adaptsegnet}                    & && 86.5 & 36.0 & 79.9 & 23.4 & 23.3 & 23.9 & 35.2 & 14.8 & 83.4 & 33.3 & 75.6 & 58.5 & 27.6 & 73.6 & 32.5 & 35.4 & 3.9  & 30.1 & 28.1 & 42.4          \\ 
MaxSquare~\cite{maxsquare}                        &&& 88.1 & 27.7 & 80.8 & 28.7 & 19.8 & 24.9 & 34.0 & 17.8 & 83.6 & 34.7 & 76.0 & 58.6 & 28.6 & 84.1 & 37.8 & 43.1 & 7.2  & 32.2 & 34.5 & 44.3          \\ 
BDL~\cite{bdl}                                    &\checkmark&\checkmark& 88.2 & 44.7 & 84.2 & 34.6 & 27.6 & 30.2 & 36.0 & 36.0 & 85.0 & 43.6 & 83.0 & 58.6 & 31.6 & 83.3 & 35.3 & 49.7 & 3.3  & 28.8 & 35.6 & 48.5          \\ 
MRNET~\cite{mrnet}                                &\checkmark&\checkmark& 90.5 & 35.0 & 84.6 & 34.3 & 24.0 & 36.8 & 44.1 & 42.7 & 84.5 & 33.6 & 82.5 & 63.1 & 34.4 & 85.8 & 32.9 & 38.2 & 2.0  & 27.1 & 41.8 & 48.3          \\ 
Stuff and things~\cite{stuffAndThings}            &\checkmark&\checkmark& 90.6 & 44.7 & 84.8 & 34.3 & 28.7 & 31.6 & 35.0 & 37.6 & 84.7 & 43.3 & 85.3 & 57.0 & 31.5 & 83.8 & 42.6 & 48.5 & 1.9  & 30.4 & 39.0 & 49.2          \\ 
FADA~\cite{fada}                                  &&\checkmark& 92.5 & 47.5 & 85.1 & 37.6 & 32.8 & 33.4 & 33.8 & 18.4 & 85.3 & 37.7 & 83.5 & 63.2 & \textbf{39.7} & \textbf{87.5} & 32.9 & 47.8 & 1.6  & 34.9 & 39.5 & 49.2          \\
LTIR~\cite{ltir}                                  &\checkmark&\checkmark& 92.9 & 55.0 & 85.3 & 34.2 & 31.1 & 34.4 & 40.8 & 34.0 & 85.2 & 40.1 & 87.1 & 61.1 & 31.1 & 82.5 & 32.3 & 42.9 & 3    & 36.4 & 46.1 & 50.2          \\   
Yang \etal~\cite{Yang_2021_WACV}                  &\checkmark&\checkmark& 91.3 & 46.0 & 84.5 & 34.4 & 29.7 & 32.6 & 35.8 & 36.4 & 84.5 & 43.2 & 83.0 & 60.0 & 32.2 & 83.2 & 35.0 & 46.7 & 0.0  & 33.7 & 42.2 & 49.2          \\
IAST~\cite{iast}                                  &&\checkmark& \textbf{93.8} & \textbf{57.8} & 85.1 &  \textbf{39.5} & 26.7 & 26.2 & 43.1 & 34.7 & 84.9 & 32.9 & 88.0 & 62.6 & 29.0 & 87.3 & 39.2 & 49.6 & \textbf{23.2} & 34.7 & 39.6 & 51.5          \\
DACS\textsuperscript{\textdagger}~\cite{dacs}     &&\checkmark& 89.9 & 39.7 & \textbf{87.9} & 30.7 & \textbf{39.5} & \textbf{38.5} & \textbf{46.4} & \textbf{52.8} & \textbf{88.0} & \textbf{44.0} & \textbf{88.8} & \textbf{67.2} & 35.8 & 84.5 & \textbf{45.7} & \textbf{50.2} & 0.0  & 27.3 & 34.0 & 52.1          \\
\hline
Ours                                              &\checkmark&\checkmark& 91.9 & 48.9 & 86.0 & 38.6 & 28.6 & 34.8 & 45.6 & 43.0 & 86.2 & 42.4 & 87.6 & 65.6 & 38.6 & 86.8 & 38.4 & 48.2 & 0.0  & \textbf{46.5} & \textbf{59.2} & \textbf{53.5}
\end{tabular}}
\end{center}
\vspace{-5mm}
\caption{Results on \gtacs{}. \textdagger{} denotes models pre-trained on  MSCOCO~\cite{coco} and ImageNet \cite{imagenet}. IT: Image Translation; ST: Self-Training.}
\label{tab:results_gta}
\end{table*}

\begin{table*}[t]
\begin{center}
\setlength{\tabcolsep}{2.5pt}
\scalebox{0.75}{
\begin{tabular}
{l|cc|cccccccccccccccc|cc}
method & IT & ST &\rotatebox{90}{Road} & \rotatebox{90}{Sidewalk} & \rotatebox{90}{Building} &\rotatebox{90}{Walls*} & \rotatebox{90}{Fence*} & \rotatebox{90}{Pole*} & \rotatebox{90}{T-light} & \rotatebox{90}{T-sign} & \rotatebox{90}{Vegetation} &  \rotatebox{90}{Sky} & \rotatebox{90}{Person} 
& \rotatebox{90}{Rider} & \rotatebox{90}{Car} & \rotatebox{90}{Bus} & \rotatebox{90}{Motorbike} & \rotatebox{90}{Bicycle} & \textbf{mIoU} & \textbf{mIoU*}   \\       
\hline
AdaptSegNet~\cite{adaptsegnet}                    &&& 84.3 & 42.7 & 77.5 & -    & -    & -    & 4.7  & 7.0  & 77.9 & 82.5 & 54.3 & 21.0 & 72.3 & 32.2 & 18.9 & 32.3 & -    & 46.7        \\ 
MaxSquare~\cite{maxsquare}                        &&& 77.4 & 34.0 & 78.7 & 5.6  & 0.2  & 27.7 & 5.8  & 9.8  & 80.7 & 83.2 & 58.5 & 20.5 & 74.1 & 32.1 & 11.0 & 29.9 & 39.3 & 45.8        \\ 
BDL~\cite{bdl}                                    &\checkmark&\checkmark& 86.0 & 46.7 & 80.3 & -    & -    & -    & 14.1 & 11.6 & 79.2 & 81.3 & 54.1 & 27.9 & 73.7 & 42.2 & 25.7 & 45.3 & -    & 51.4        \\ 
MRNET~\cite{mrnet}                                &\checkmark&\checkmark& 83.1 & 38.2 & 81.7 & 9.3  & 1.0  & 35.1 & 30.3 & 19.9 & 82.0 & 80.1 & 62.8 & 21.1 & 84.4 & 37.8 & 24.5 & \textbf{53.3} & 46.5 & 53.8        \\ 
Stuff and things~\cite{stuffAndThings}            &\checkmark&\checkmark& 83.0 & 44.0 & 80.3 & -    & -    & -    & 17.1 & 15.8 & 80.5 & 81.8 & 59.9 & 33.1 & 70.2 & 37.3 & 28.5 & 45.8 & -    & 52.1        \\ 
FADA~\cite{fada}                                  &&\checkmark& 84.5 & 40.1 & 83.1 & 4.8  & 0.0  & 34.3 & 20.1 & 27.2 & 84.8 & 84.0 & 53.5 & 22.6 & 85.4 & 43.7 & 26.8 & 27.8 & 45.2 & 52.5        \\
LTIR~\cite{ltir}                                  &\checkmark&\checkmark& \textbf{92.6} & 53.2 & 79.2 & -    & -    & -    & 1.6  & 7.5  & 78.6 & 84.4 & 52.6 & 20.0 & 82.1 & 34.8 & 14.6 & 39.4 & -    & 49.3        \\
Yang \etal~\cite{Yang_2021_WACV}                  &\checkmark&\checkmark& 82.5 & 42.2 & 81.3 & -    & -    & -    & 18.3 & 15.9 & 80.6 & 83.5 & 61.4 & 33.2 & 72.9 & 39.3 & 26.6 & 43.9 & -    & 52.4        \\
IAST~\cite{iast}                                  &&\checkmark& 81.9 & 41.5 & 83.3 & 17.7 & \textbf{4.6}  & 32.3 & \textbf{30.9} & 28.8 & 83.4 & 85.0 & 65.5 & 30.8 & \textbf{86.5} & 38.2 & \textbf{33.1} & 52.7 & \textbf{49.8} & \textbf{57.0}        \\
DACS\textsuperscript{\textdagger}~\cite{dacs}       &&\checkmark& 80.6 & 25.1 & 81.9 & \textbf{21.5} & 2.6  & \textbf{37.2} & 22.7 & 24.0 & 83.7 & \textbf{90.8} & \textbf{67.6} & \textbf{38.3} & 82.9 & 38.9 & 28.5 & 47.6 & 48.3 & 54.8        \\
\hline
Ours                                              &\checkmark&\checkmark& 90.4 & \textbf{51.1} & \textbf{83.4} & 3.0 & 0.0  & 32.3 & 25.3 & \textbf{31.0} & \textbf{84.8} & 85.5 & 59.3 & 30.1 & 82.6 & \textbf{53.2} & 17.5 & 45.6 & 48.4 & 56.9
\end{tabular}}
\end{center}
\vspace{-5mm}
\caption{Results on \synthiacs{}. \textdagger{} denotes models pre-trained with MSCOCO~\cite{coco} and ImageNet \cite{imagenet}. IT: Image Translation; ST: Self-Training. The 13 classes with $^*$ are used to compute mIoU$^*$.}
\label{tab:results_synthia}
\end{table*}

\section{Implementation}

\subsection{Architecture}
According to standard practice in UDA for semantic segmentation \cite{adaptsegnet, maxsquare, bdl, mrnet, stuffAndThings, fada, ltir}, we deploy the Deeplab-v2~\cite{deeplabv2} architecture, with a dilated ResNet101 pre-trained on ImageNet~\cite{imagenet} and output stride 8. The ASPP~\cite{deeplabv2} module acts as classifier. 
We use this architecture for both the initialization step and the distillation step. For more details on the additional modules of the adaptation step we refer to the supplementary material.

\subsection{Training Details}
Our pipeline is implemented in PyTorch~\cite{pytorch} and trained on a single NVIDIA 2080Ti GPU with 12GB of memory. 
We train for 20 epochs in the first two steps, while we set the number of epochs to 25 for the final distillation with batch size 4 in all cases.
We use random scaling, random cropping at $1024\times892$, and color jittering in our data augmentation pipeline.
Akin to  previous works, we freeze Batch-Normalization layers \cite{batchnorm} while performing the initialization and adaptation step. For the last step, instead, we activate these layers.
We adopt the One Cycle learning rate policy~\cite{onecycle} for each training, with maximum learning rate $10^{-3}$ and SGD as optimizer.

\section{Experiments}

\begin{table*}[t]
\begin{center}
\setlength{\tabcolsep}{0.9mm}
\scalebox{0.76}{
\begin{tabular}
{l|c|c|ccccccccccccc|cc}
				City & Method & ST & \rotatebox{90}{road} & \rotatebox{90}{sidewalk} & \rotatebox{90}{building} & \rotatebox{90}{light} & \rotatebox{90}{sign} & \rotatebox{90}{veg.} & \rotatebox{90}{sky} & \rotatebox{90}{person} & \rotatebox{90}{rider} & \rotatebox{90}{car} & \rotatebox{90}{bus} & \rotatebox{90}{motor} & \rotatebox{90}{bike} & mIoU (\%) \\
				\hline
				\multirow{5}{*}{Rome} 
				& Source only                     & & 85.9 & 40.0 & 86.0 & 9.0  & 25.4 & 82.4 & 90.5 & 38.8 & 25.9 & 81.6 & 52.0 & 48.7 & 6.7  & 51.9 \\
				& CBST~\cite{cbst}                & \checkmark& 87.1 & 43.9 & \textbf{89.7} & 14.8 & 47.7 & 85.4 & 90.3 & 45.4 & 26.6 & \textbf{85.4} & 20.5 & 49.8 & 10.3 & 53.6 \\
				& AdaptSegNet~\cite{adaptsegnet}  & & 83.9 & 34.2 & 88.3 & 18.8 & 40.2 & \textbf{86.2} & \textbf{93.1} & 47.8 & 21.7 & 80.9 & 47.8 & 48.3 & 8.6  & 53.8 \\
				& MaxSquare~\cite{maxsquare}      & & 80.0 & 27.6 & 87.0 & \textbf{20.8} & \textbf{42.5} & 85.1 & 92.4 & 46.7 & 22.9 & 82.1 & 53.5 & 50.8 & 8.8  & 53.9 \\
				& FADA~\cite{fada}                & \checkmark & 84.9 & 35.8 & 88.3 & 20.5 & 40.1 & 85.9 & 92.8 & \textbf{56.2} & 23.2 & 83.6 & 31.8 & 53.2 & \textbf{14.6} & 54.7 \\
				& Ours                            & \checkmark & \textbf{89.4} & \textbf{48.2} & 87.5 &  \textbf{26.3}  & 37.2 & 83.1 & 90.7 & \textbf{55.2} & \textbf{42.1} & 84.8 & \textbf{66.6} & \textbf{59.2} & 11.1 & \textbf{60.1} \\
				\hline
				\multirow{5}{*}{Rio}
				& Source only                     & & 80.4 & 53.8 & 80.7 & 4.0  & 10.9 & 74.4 & 87.8 & 48.5 & 25.0 & 72.1 & 36.1 & 30.2 & 12.5 & 47.4 \\
				& CBST~\cite{cbst}                & \checkmark& 84.3 & 55.2 & 85.4 & 19.6 & \textbf{30.1} & 80.5 & 77.9 & 55.2 & 28.6 & \textbf{79.7} & 33.2 & 37.6 & 11.5 & 52.2 \\
				& AdaptSegNet~\cite{adaptsegnet}  & & 76.2 & 44.7 & 84.6 & 9.3  & 25.5 & \textbf{81.8} & 87.3 & 55.3 & 32.7 & 74.3 & 28.9 & 43.0 & 27.6 & 51.6\\
				& MaxSquare~\cite{maxsquare}      & & 70.9 & 39.2 & \textbf{85.6} & \textbf{14.5} & 19.7 & \textbf{81.8} & 88.1 & 55.2 & 31.5 & 77.2 & 39.3 & 43.1 & 30.1 & 52.0 \\
			    & FADA~\cite{fada}                & \checkmark & 80.6 & 53.4 & 84.2 & 5.8  & 23.0 & 78.4 & 87.7 & \textbf{60.2} & 26.4 & 77.1 & 37.6 & \textbf{53.7} & \textbf{42.3} & 54.7  \\
				& Ours                            &\checkmark & \textbf{86.6} & \textbf{63.3} & 82.3 & 10.3 & 19.8 & 73.9 & \textbf{88.4} & 57.5 & \textbf{41.3} & 78.1 & \textbf{51.5} & 40.0 & 19.4 & \textbf{54.8} \\
				\hline
				\multirow{5}{*}{Tokyo} 
				& Source only                    & & 86.0 & 38.8 & 76.6 & 11.7 & 12.3 & 80.0 & 89.5 & 44.9 & 28.0 & 71.5 & 4.7 & 27.1 & 42.2 & 47.2 \\
				& CBST~\cite{cbst}               & \checkmark& 85.2 & 33.6 & \textbf{80.4} & 8.3 & \textbf{31.1} & 83.9 & 78.2 & 53.2 & 28.9 & 72.7 & 4.4 & 27.0 & 47.0 & 48.8 \\
				& AdaptSegNet~\cite{adaptsegnet} & & 81.5 & 26.0 & 77.8 & 17.8 & 26.8 & 82.7 & 90.9 & 55.8 & \textbf{38.0} & 72.1 & 4.2 & 24.5 & 50.8 & 49.9\\
				& MaxSquare~\cite{maxsquare}     & & 79.3 & 28.5 & 78.3 & 14.5 & 27.9 & 82.8 & 89.6 & 57.3 & 31.9 & 71.9 & 6.0 & 29.1 & 49.2 & 49.7 \\
			    & FADA~\cite{fada}               & \checkmark & 85.8 & 39.5 & 76.0 & 14.7 & 24.9 & \textbf{84.6} & \textbf{91.7} & 62.2 & 27.7 & 71.4 & 3.0 & 29.3 & \textbf{56.3} & 51.3 \\
			    & Ours                           & \checkmark & \textbf{87.8} & \textbf{41.0} & 79.6 & \textbf{20.3} & 24.2 & 80.2 & 90.0 & \textbf{62.3} & 30.8 & \textbf{74.0} & \textbf{6.4} & \textbf{32.7} & 50.0 & \textbf{52.4} \\
				\hline
				\multirow{5}{*}{Taipei} 
				& Source only                    & & 85.0 & 38.1 & 82.2 & 17.8 & 8.9  & 75.2 & 91.4 & 23.9 & 19.6 & 69.2 & 45.9 & 49.4 & 16.0 & 47.9 \\
				& CBST~\cite{cbst}               &\checkmark & 86.1 & 35.2 & 84.2 & 15.0 & 22.2 & 75.6 & 74.9 & 22.7 & 33.1 & 78.0 & 37.6 & 58.0 & 30.9 & 50.3 \\
				& AdaptSegNet~\cite{adaptsegnet} & & 81.7 & 29.5 & 85.2 & 26.4 & 15.6 & 76.7 & 91.7 & 31.0 & 12.5 & 71.5 & 41.1 & 47.3 & 27.7 & 49.1 \\
				& MaxSquare~\cite{maxsquare}     & & 81.2 & 32.8 & 85.4 & 31.9 & 14.7 & 78.3 & 92.7 & 28.3 & 8.6  & 68.2 & 42.2 & 51.3 & 32.4 & 49.8 \\
				& FADA~\cite{fada}               & \checkmark& 86.0 & 42.3 & 86.1 & 6.2  & 20.5 & 78.3 & 92.7 & 47.2 & 17.7 & 72.2 & 37.2 & 54.3 & 44.0 & 52.7 \\
				& Ours                           & \checkmark & \textbf{95.6}  & \textbf{78.9} & \textbf{94.3} & \textbf{45.9} & \textbf{70.3} & \textbf{93.0} & \textbf{96.2} & \textbf{63.3} & \textbf{51.3} & \textbf{90.5} & \textbf{83.6} & \textbf{84.8} & \textbf{56.5} & \textbf{55.7} \\
				\hline
\end{tabular}}
\end{center}
\caption{Results for the Cross-City experiments. ST: Self-Training.}
\label{tab:crosscity}
\end{table*}

\subsection{Datasets}
We test our method on both synthetic-to-real and real-to-real adaptation. We set GTA \cite{gta} or SYNTHIA \cite{synthia} as source datasets and Cityscapes \cite{Cityscapes} as target for the former setting, while we use Cityscapes as source and the NTHU \cite{NTHU} dataset as target for the latter.
GTA5 is a synthetic dataset that contains 24,966 annotated images of $1914 \times 1052$ resolution. As for SYNTHIA, we use the SYNTHIA-RAND-CITYSCAPES subset, which is a collection of 9,400 synthetic images with resolution $1280 \times 760$.
The Cityscapes dataset is a high-quality collection of real images of $2048 \times 1024$ resolution. The dataset has 2975 and 500 images for the training and validation split, respectively.
For the synthetic-to-real case, we only utilize the training split without labels for training, and test on the validation set as done in previous works \cite{adaptsegnet, cbst, bdl}.
The NTHU dataset is a collection of images taken from four different cities with $2048 \times 1024$ resolution: Rio, Rome, Tokyo, and Taipei. For each city, 3200 unlabeled images are available for the adaptation phase, and 100 labeled images for the  evaluation. 
For fair comparison to other models, we compute the mIoU by considering all 19 classes in the \gtacs{} benchmark, 16 or 13 shared classes for \synthiacs{}, and 13 common classes for the cross-city adaptation setting.

\subsection{Synthetic-to-real adaptation}
To test our framework, we follow standard practice \cite{adaptsegnet, cbst, bdl, mrnet, advent, maxsquare} and report the results for the synthetic-to-real adaptation in the \gtacs{} and \synthiacs{} benchmarks in \cref{tab:results_gta} and \cref{tab:results_synthia} respectively.
We obtain state-of-the-art performance in the former setting, surpassing also recent methods such as \cite{iast} that performs many iterations of self-training.
We also improve over \cite{dacs} for \gtacs{}, which, differently from all other methods, pre-trains the baseline network not only on ImageNet\cite{imagenet} but also on MSCOCO\cite{coco}. We argue that pre-training on more tasks and real annotated data notably improves the baseline performance of the synthetic-to-real benchmark.
For \gtacs{}, we note that, thanks to our low-level adaptation, we can boost performances for fine-detailed classes such as \textit{Bicycle} and \textit{Motorcycle}.
Regarding \synthiacs{}, we obtain competitive performance, showing that our method can work also in this challenging scenario in which the source synthetic domain exhibits many bird's-eye views that are very different from the one in Cityscapes. Indeed our method is only slightly inferior to IAST\cite{iast} and again superior to  \cite{dacs} that performs a similar data augmentation.

\begin{table}[t!]
    \centering
    \small
    \scalebox{0.8}{
    \begin{tabular}{c|ccccc|c|c}
        & & & & & & \cellcolor{YellowOrange}GTA & \cellcolor{blue!25}Synthia \\
        \hline
         \textbf{Step} & IT & ST & A & W & D & mIoU & mIoU  \\
         \hline
         Initialization &\checkmark&  &&&$\mathcal{S}$& 47.3 & 41.6 \\
         \hline
          \multirow{3}{*}{Adaptation} &\checkmark&\checkmark&&&$\mathcal{S,T}$& 49.8 & 43.5 \\
                                      &\checkmark&\checkmark&\checkmark&&$\mathcal{S,T}$& 52.0 & 46.4 \\                                      &\checkmark&\checkmark&\checkmark&\checkmark&$\mathcal{S,T}$& 52.6 & 46.9\\
         \hline
          Distillation &\checkmark&\checkmark&\checkmark&&$\mathcal{T}$& 53.5 & 48.4\\
         \hline
          Oracle &&&&&$\mathcal{T}$& 63.8 & 65.1\\
    \end{tabular}}
    \caption{Ablation studies on \gtacs{} (second-to-last) and \synthiacs{} (last) columns. IT: image translation; ST: Self-Training; W: low-level adaptation; A: Data Augmentation; D: Training domain.}
    \label{tab:ablation_modules}
\end{table}


\subsection{Cross-city adaptation}
We report in \cref{tab:crosscity} our performance for the real-to-real setting. Our proposal shows great results, confirming the generalization properties of our contributions on diverse settings. We improve performance with respect to previous works for all the cities. Our model achieves 60\% in mIoU in Rome, which is likely the most similar to the German cities used in the Cityscapes dataset. Nonetheless, we achieve strong results even for more distant domains, e.g. as in the case of Taipei, improving by 7.8\% with respect to the model trained only on the source domain. For the Cross-city adaptation setting, differently from the other settings, we use images of both domains in our \textit{distillation} step to exploit the perfect annotations available in the similar source domain.

\subsection{Ablation Studies}

In this section, we analyze the contribution provided by each component of our framework and motivate our design choices. In \cref{tab:ablation_modules} we detail the results for both \gtacs{} and \synthiacs{}. The first row reports the performance obtained using only translated source domain images. This is nowadays a common building block of many UDA frameworks, and we also consider it our baseline on which we build our pipeline.
In the adaptation section instead, we isolate both our contributions and use the model trained in the initialization step to extract pseudo-labels for the target domain as explained in \cref{sec:selftraining} and train on both domains simultaneously.
When applying a naive self-training strategy (i.e. training directly on pseudo-labels) we already obtain a significant boost (+2.5\% and +1.9\%) respectively. However, when deploying the proposed data augmentation (row 3), we observe an even greater boost: +4.8\% for both settings. This clearly demonstrates the effectiveness of our data augmentation and its applicability to diverse scenarios.
Then, applying the proposed low-level adaptation (row 4) also yields an additional contribution overall: about +0.6\% on top of the data augmentation version. We argue that is noticeable, especially when performances are already high, as in our case, and the strongest competitors are all within a narrow window. 
Finally, in row 5, we distill our full model (i.e. row 4) into a simple Deeplab-v2 for efficient inference time and apply once again the proposed data augmentation.
Remarkably, this further improves performance with respect to the distilled model and avoids the typical pseduo-labels overfitting behavior when employing many steps of self-training.

Moreover, to motivate our intuition that shallow features are amenable to guide the warping process, we compare the results obtained by applying our adaptation step in the \gtacs{} setting at the three different levels of the backbone before the last module achieving 52.6\%, 51.6\%, and 51.8\% mIoU for layers \textit{Conv1}, \textit{Conv2}, and \textit{Conv3} respectively. Thus, the best result is achieved by using the first convolutional block of the architecture, while on \textit{Conv2} and \textit{Conv3} results are comparable (see \cref{fig:network} for layer names).

\subsection{Performance Along Class Boundaries}

\begin{figure}[h]
    \centering
    \includegraphics[width=0.65\linewidth]{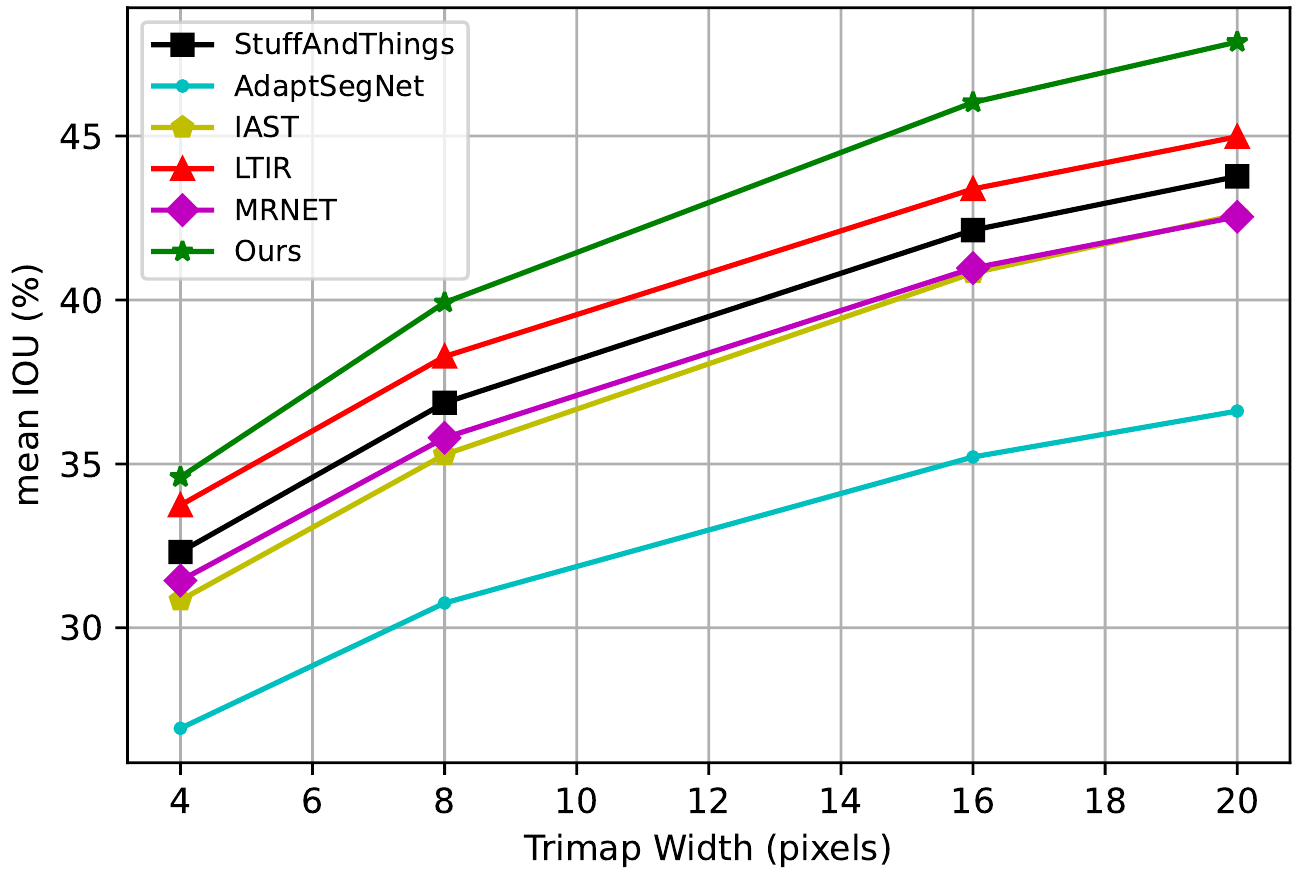}
    \caption{mIOU on \gtacs{} as a function of trimap band width around class boundaries.}
    \label{fig:trimap}
\end{figure}

\begin{figure}[h]
    \centering
    \includegraphics[width=0.75\linewidth]{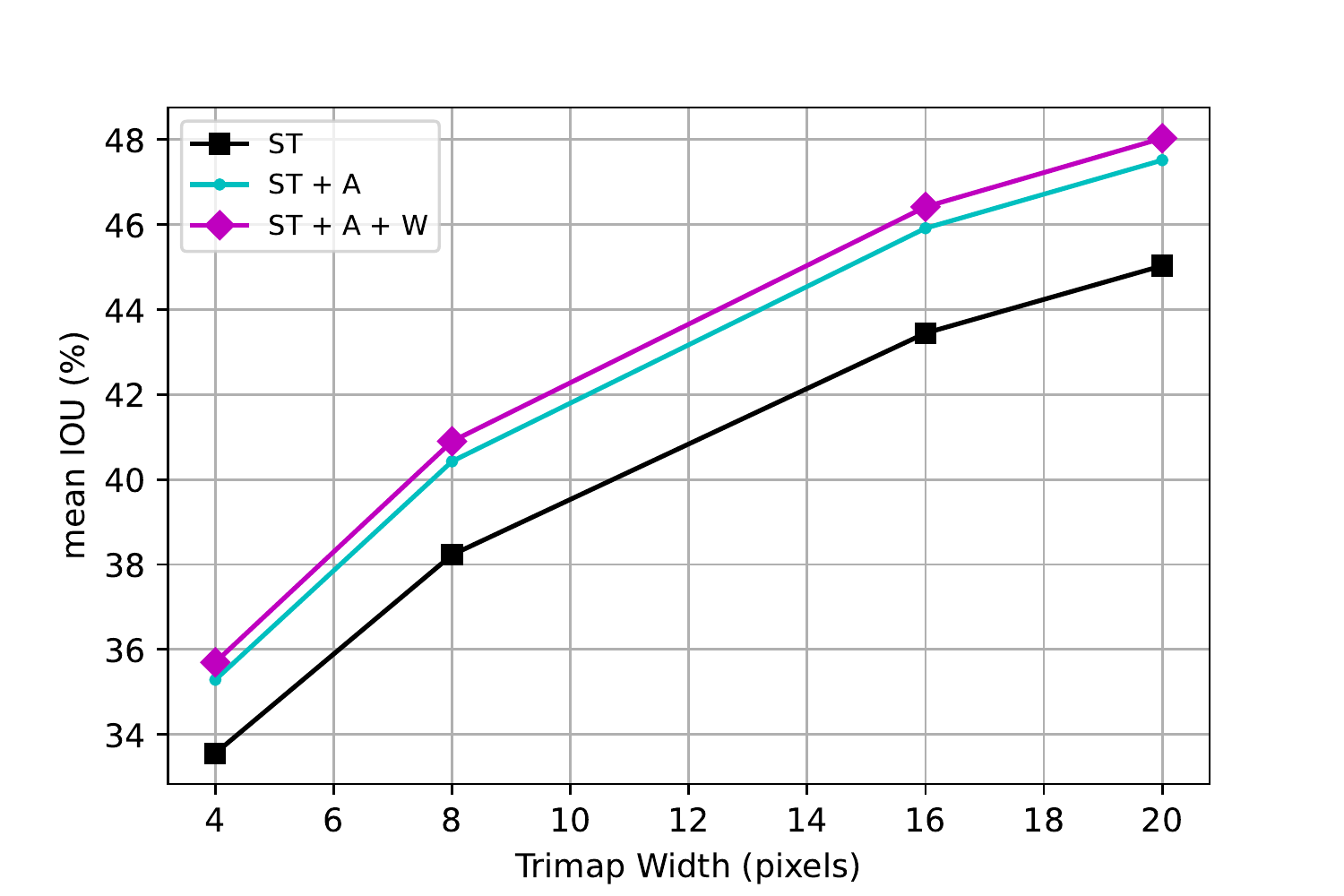}
    \caption{mIOU on \gtacs{} as a function of trimap band width around class boundaries. We report results for the three versions of the \textit{adaptation} step of \cref{tab:ablation_modules}.}
    \label{fig:trimap_warp}
\end{figure}

In this section, we test the segmentation accuracy with the trimap experiment \cite{chen2018encoder, krahenbuhl2011efficient, chen2017deeplab, kohli2009robust} to quantify the accuracy of the proposed method along semantic edges.
Specifically, we evaluate in terms of mIoU pixels within four bandwidths (4, 8, 16, 20 pixels) around class boundaries (trimaps). 
We first compare our final model against other frameworks in  \cref{fig:trimap}. We observe that our method is more accurate w.r.t. all other competitors in all the tested bandwidths, validating our main goal that is improving precision along class boundaries. We also highlight that although the green line is obtained from a distilled model (row 5 of \cref{tab:ablation_modules}), that does not include the additional modules presented in \cref{sec:lowalign}, it is still able to maintain strong performances at semantic boundaries thanks to the precise pseudo-labels extracted from the adaptation step.
We refer to supplementary materials for some qualitative examples. Then, we assess in \cref{fig:trimap_warp} how our contributions affect performances on semantic boundaries.
To this end, we repeat the same trimap experiment using the intermediate steps of our pipeline i.e. row 2, 3, and 4 of  \cref{tab:ablation_modules}. When applying all our contributions (purple line), we are able to improve by a large margin over the self-training strategy (black line) confirming that the additional modules account for an improvement along semantic edges. Furthermore, activating the low-level adaptation strategy maintains its improvements along semantic edges over the data augmentation only version (cyan line), leading to better pseudo-labels for the distillation step.

\subsection{Comparison with other data augmentations}
We compare our data augmentation, one of our main contributions, with the one introduced in \cite{dacs}.
More specifically, we apply this data augmentation in the adaptation step as in row 3 of \cref{tab:ablation_modules}, i.e. without the low-level adaptation modules to isolate the data-augmentation effect. We augment target images randomly pasting objects from the source domain,
using the open source implementation of \cite{dacs}. With this strategy, we only obtain 51.0\% in terms of mIoU, while with our technique the mIoU raises to 52.2\%, confirming our intuition that looking for target instances is more effective than forcing the network to identify source objects as done \cite{dacs} during the self-training step.

\subsection{Displacement map visualization}
\begin{figure}[t]
    \centering
    \includegraphics[scale=0.21]{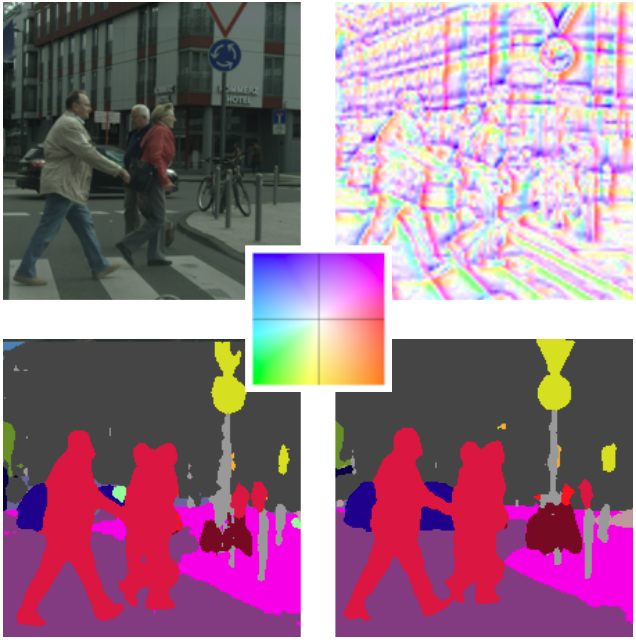}
    \caption{Top left: input target image. Top right: estimated 2D displacement. Bottom left: semantic map from a model trained on translated images. Bottom Right: Our results, improved on class boundaries by using the warping module.
    Colors and lightness in the middle indicates the warping direction with the corresponding intensity.
    }
    \label{fig:flow}
\end{figure}

In this section, we analyze the displacement map learned by the model. As \cref{fig:flow} shows, the 2D map that guides the warping process is consistent with our intuition that the displacement is more pronounced at the boundaries, while areas within regions such as the body of a person, are characterized by a low displacement (i.e. white color).
Moreover, we can appreciate that when the warping is applied according to the estimated displacement field (top-right), 
the contours of small objects such as poles, traffic signs, and persons are better delineated (bottom-right). On the other hand, in the bottom-left mask, these objects are coarsely segmented when using a segmentation model train with translated images only. We also highlight that the displacement field is agnostic to semantic class (it only considers boundaries), and even though it captures other kinds of edges (i.e. not only semantic ones), it leads to computing an average of patches belonging to the same class.

\section{Conclusion}
In this paper, we have proposed a novel framework for UDA for semantic segmentation that explicitly focuses on improving accuracy along class boundaries. 
We have shown that we can exploit domain-invariant shallow features to estimate a displacement map used to achieve sharp predictions along semantic edges. 
Jointly with a novel data augmentation technique that preserves fine edge information during self-training, our approach achieves better accuracy along class boundaries w.r.t. previous methods.
%

{\small
\bibliographystyle{ieee_fullname}
\bibliography{egbib}
}

\newpage\phantom{Supplementary}
\multido{\i=1+1}{6}{


\section*{Supplementary material}

# Supplementary Material for Shallow Features Guide Unsupervised Domain Adaptation for Semantic Segmentation at Class Boundaries

Adriano Cardace    Pierluigi Zama Ramirez    Samuele Salti    Luigi Di Stefano

Department of Computer Science and Engineering (DISI)

University of Bologna, Italy

{adriano.cardace2, pierluigi.zama}@unibo.it

| Layer | K | In/Out | Input |
|---|---|---|---|
| **Edge Module** | | | |
| Conv1 + BN + ReLU | 3 | 256/128 | input |
| Conv2 + BN + ReLU | 3 | 128/64 | Conv1 |
| Conv3 | 1 | 64/1 | Conv2 |
| **Warping Module** | | | |
| Conv1 + BN + ReLU | 3 | 256/256 | input |
| Conv2 + BN + ReLU | 3 | 256/256 | Conv1 |
| Conv3 | 1 | 256/2 | Conv2 |
| **Final Classifier** | | | |
| Conv1 + BN + ReLU | 3 | 128/128 | input |
| Conv2 + BN + ReLU | 3 | 128/64 | Conv1 |
| Conv3 | 1 | 64/C | Conv2 |

Table 1: Detailed structure of the additional components used in the *adaptation* step.

## 1. Network Architecture

In this section, we provide a more detailed description of our architecture used for the *adaptation* step. We rely on the widely adopted Deeplab-v2 backbone and feed the output of *conv2* to two additional blocks to predict semantic edges and estimate the 2D displacement grid. Moreover, as explained in Sec. 3 of the main paper, a third component that acts as a classifier is introduced to predict the final segmentation mask given the fine-grained feature map. Table 1 reports the architectural details of the three additional components. For each layer, we report kernel size (K) and number of input/output channels.

## 2. Training details

We provide here some additional details for each experiment.

### 2.1. Synthetic-to-real adaptation

As explained in Sec. 3.2 of the main paper, we exclude certain classes in the data augmentation pipeline to avoid scene occlusion in the newly generated images. Indeed, classes such as *road* and *building* typically cover a large portion of the input images and are more unlikely to create plausible scenes when pasted into source images. For this reason, we exclude *road, sidewalk, building, vegetation, sky* for both GTA5→Cityscapes and SYNTHIA→Cityscapes. At inference time, we follow common practice [4, 7, 1, 6], and resize the input image to $512 \times 1024$ pixels, while the final prediction is bi-linearly up-sampled to $1024 \times 2048$ pixels to compute the mIoU score based on the provided annotation.

### 2.2. Real-to-real adaptation

In the real-to-real adaptation scenario, we train on Cityscapes [3] and test on the four different cities of the NTHU [2] dataset. Since the source domain contains only 2975 training pairs, we crop images at $512 \times 1024$. Following standard practise [7, 1, 5], we merge *pole, fence, wall* into the class *buildings*, *truck* into *car* and *terrain* into *vegetation*. Akin to the synthetic-to-real setting, at test time we resize the input image to $512 \times 1024$ pixels and up-sample the prediction to the same resolution of the ground-truth, i.e. $1024 \times 2048$ pixels.

## 3. Qualitative Results

In this section, we provide additional qualitative results for the three benchmarks in which we tested our method. Fig. 1 shows predictions obtained for the GTA5→Cityscapes benchmark, while Fig. 2 and Fig. 3 deal with SYNTHIA→Cityscapes and Cityscapes→CrossCity, respectively. We observe that in all cases our method achieves excellent performance along class boundaries. In particular, our method can yield very precise segmentation masks for objects such as *pole*, *persons* and *traffic sign*.

## 4. Displacement Maps

We highlight in Fig. 4 the effect of our warping module, qualitatively proving the intuition the by means of the

displacement map we can obtain sharp segmentation masks in the target domain as well. In Fig. 4 (a), for example, it is possible to appreciate how the legs of the rider (bottom-right) are less noisy compared to those obtained by a model trained on translated images only (bottom-left). Similarly, Fig. 4 (d) shows how the correct shape of the traffic sign is recovered thanks to the displacement map.

## 5. Data Augmentation

In this section, we show several training pairs generated with our data augmentation pipeline, described in Sec. 3.3 of the main paper. Fig. 5 and Fig. 6 deal with GTA5→Cityscapes and SYNTHIA→Cityscapes, respectively. It is worth noticing that we perform our data augmentation pipeline on translated source images (left-side). Indeed, these images resemble in color the gray-stylish images typical of Cityscapes[3]. In the top row of Fig. 5 for example, we can appreciate how the class fence, which is correctly classified (see predictions on the right-side), is pasted into a translated source image on both sides of the road. Similarly, in the top row of Fig. 6, an instance of the class *bus* is correctly classified and inserted into the source scene.

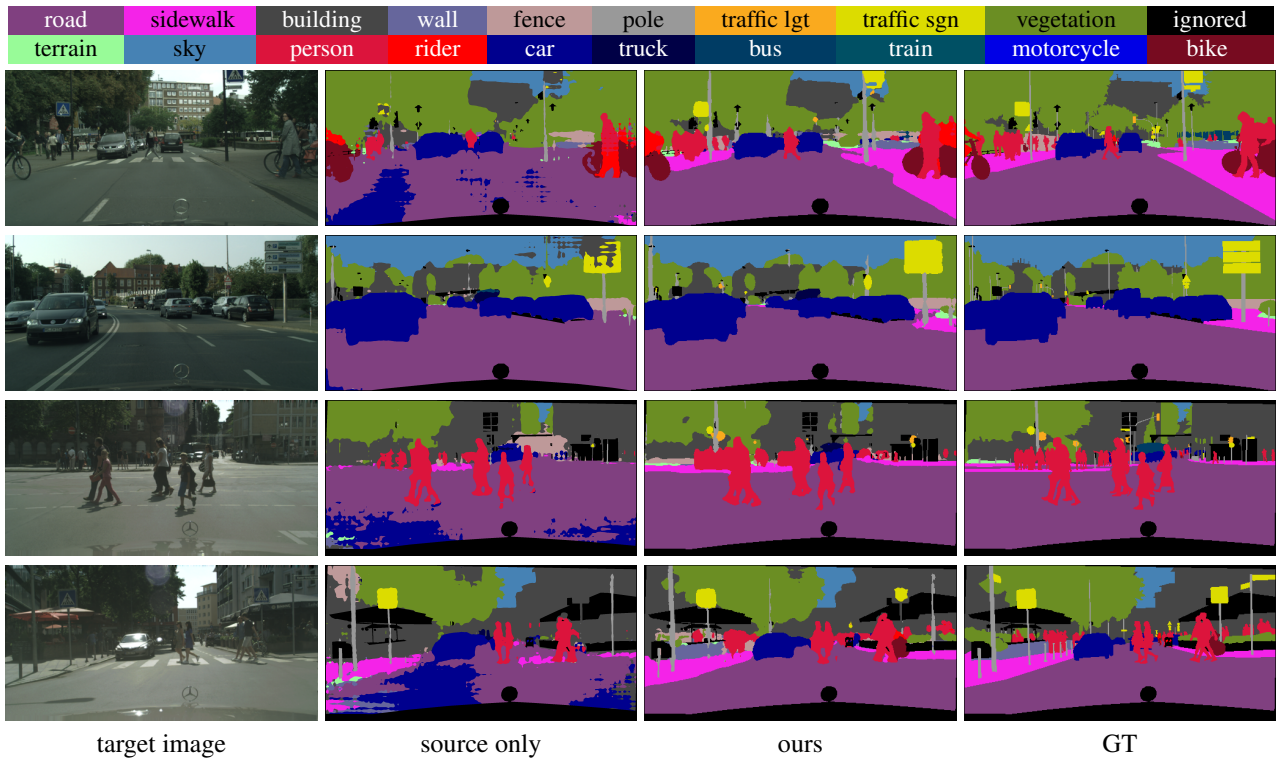

Figure 1: Qualitative results on GTA5→Cityscapes. From left to right: input image, source only, our method, ground-truth.

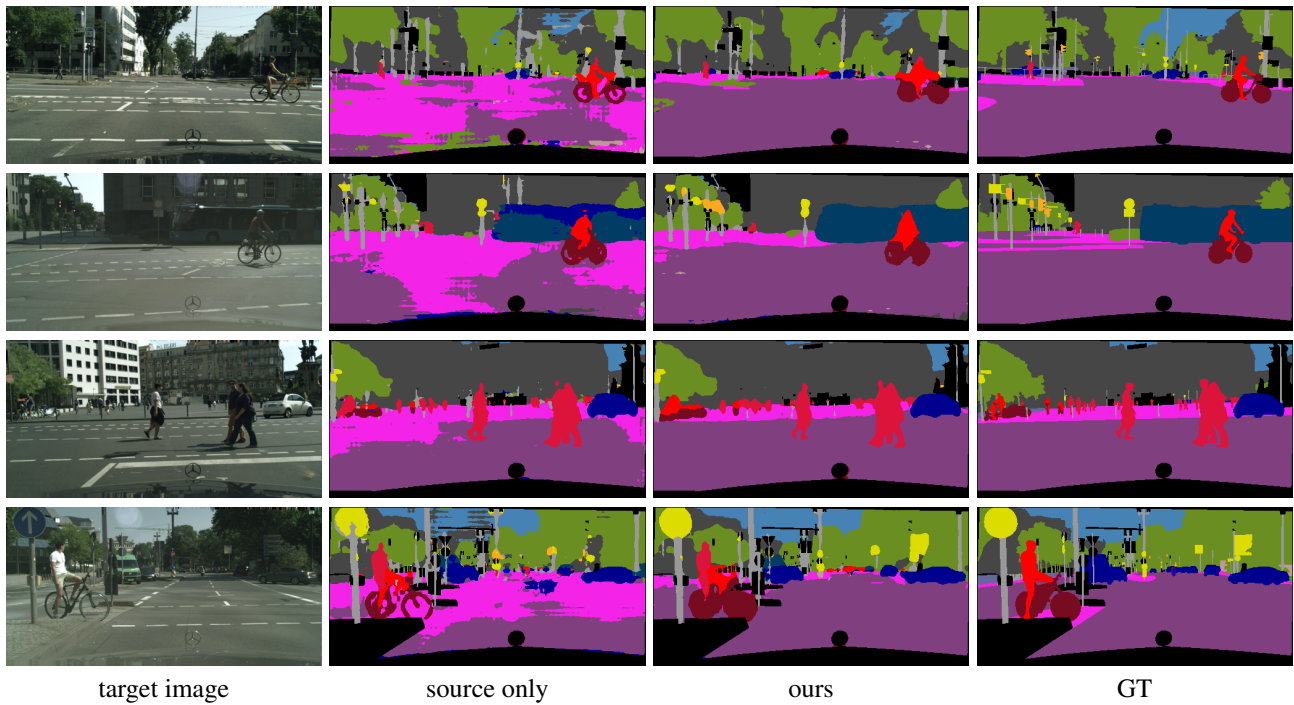

Figure 2: Qualitative results on SYNTHIA→Cityscapes. From left to right: input image, source only, our method.

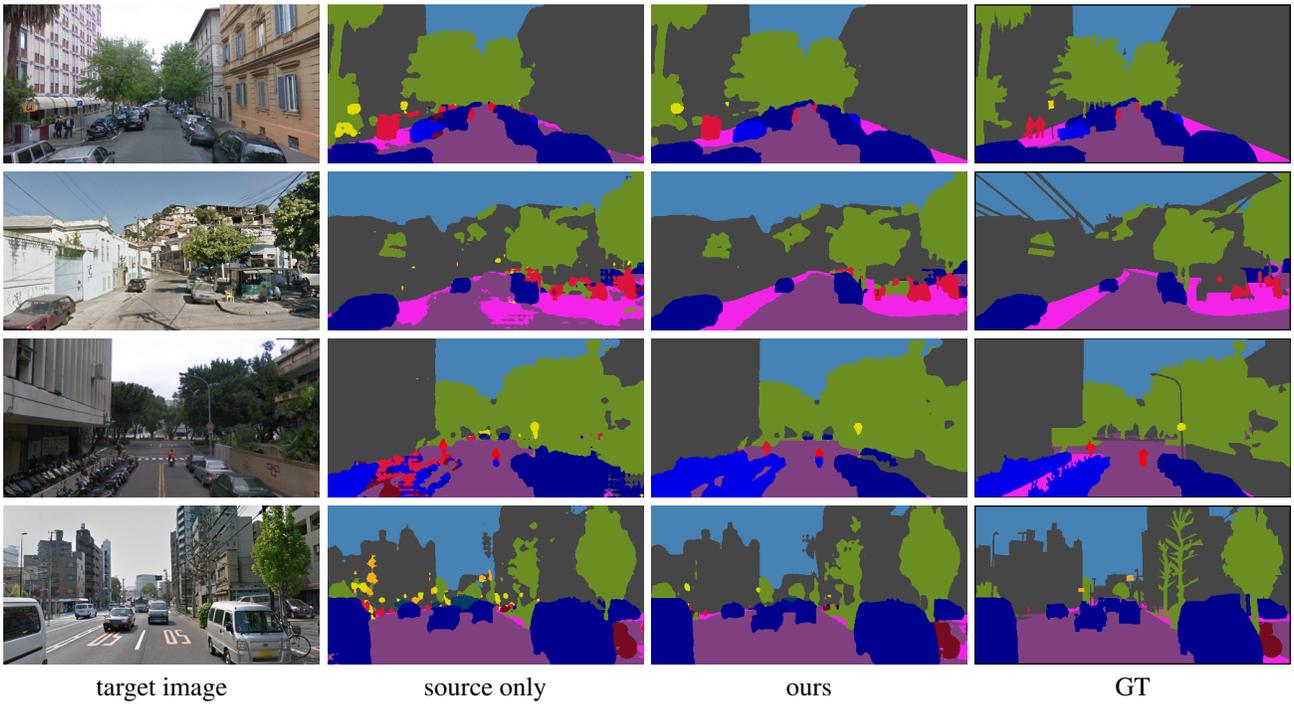

Figure 3: Qualitative results on Cityscapes→CrossCity. From left to right: input image, source only, our method. From top to bottom: Rome, Rio, Taipei, Tokyo.

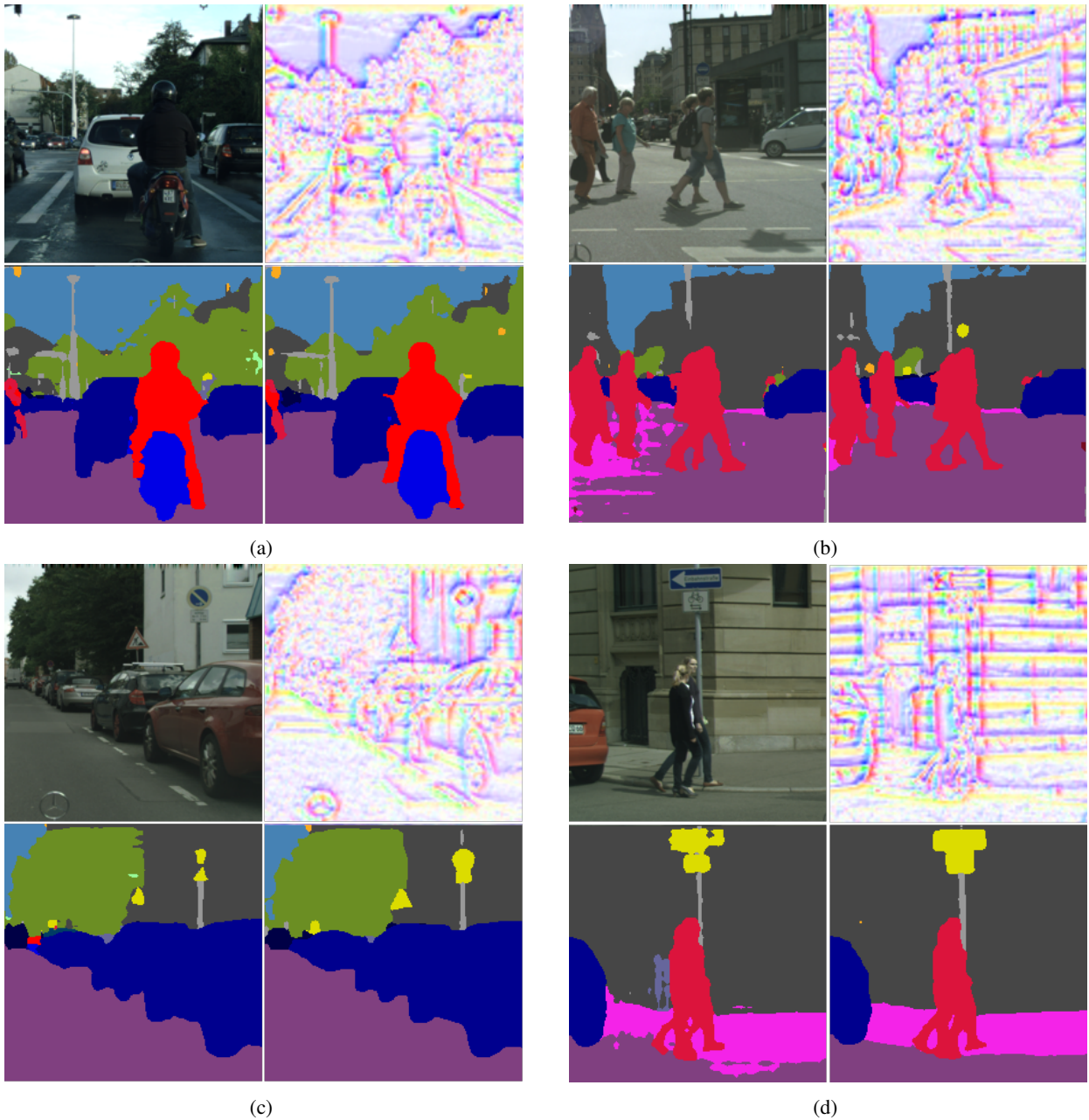

Figure 4: Displacement field visualization in GTA5→Cityscapes. Top left: input image. Top right: estimated displacement field. Bottom left: semantic map from a model trained on translated images. Bottom Right: Our results, providing more accurate segmentation along class boundaries thanks to feature warping based on the estimated displacement field.

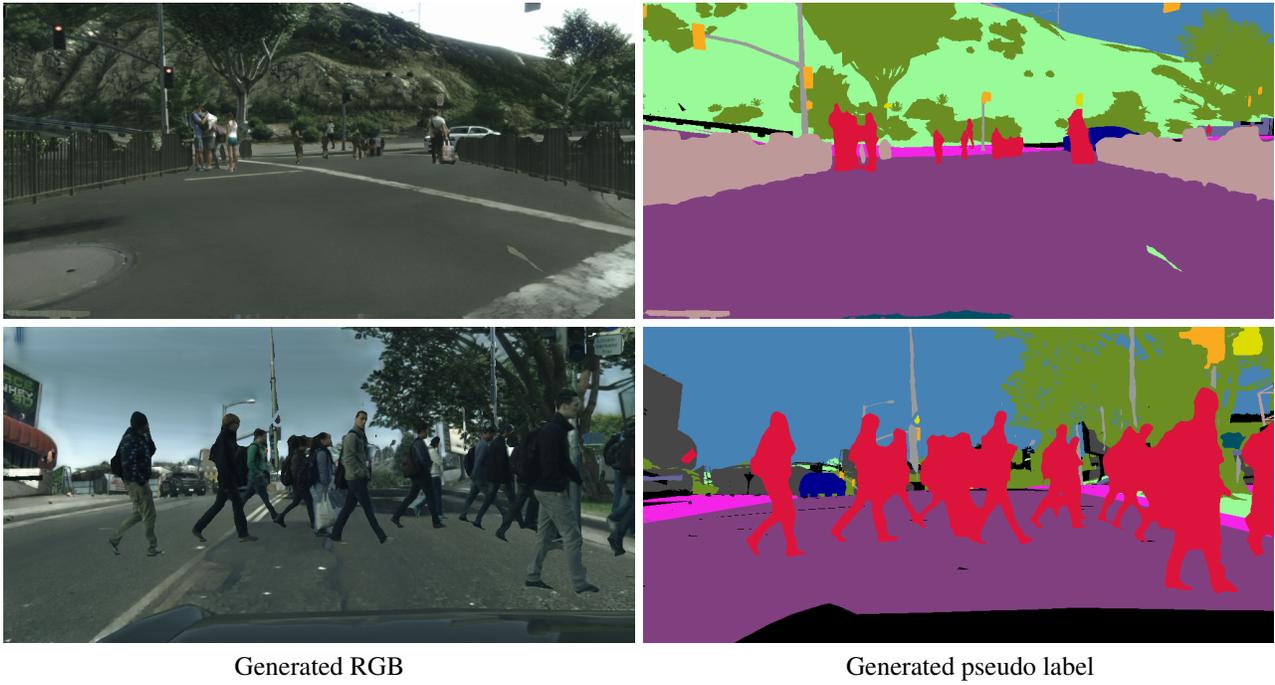

Generated RGB | Generated pseudo label

Figure 5: Examples of two newly generated training pairs with our data augmentation for GTA5→Cityscapes. Left images consist of translated source images enhanced with target objects, while right images represent the associated pseudo-labels.

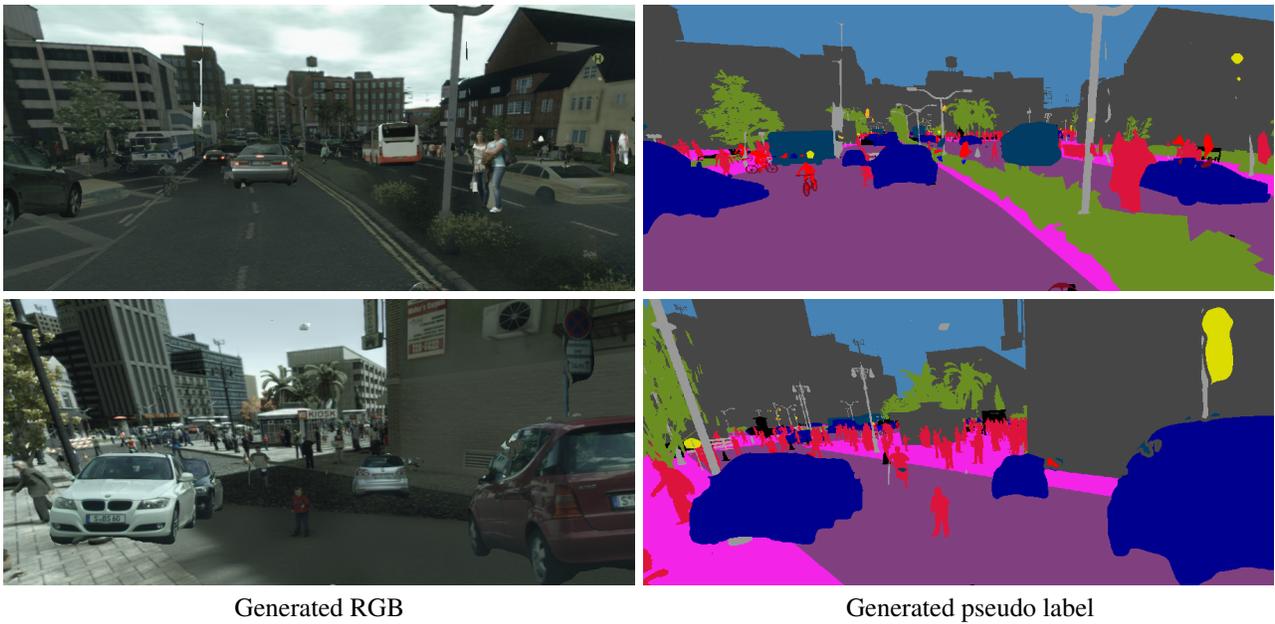

Generated RGB | Generated pseudo label

Figure 6: Examples of two newly generated training pairs with our data augmentation for SYNTHIA→Cityscapes. Left images consist of translated source images enhanced with target objects, while right images represent the associated pseudo-labels.


}

\end{document}